\documentclass[10pt,twocolumn,letterpaper]{article}

\usepackage[pagenumbers]{cvpr} 

\usepackage{multirow}

\definecolor{cvprblue}{rgb}{0.21,0.49,0.74}
\usepackage[pagebackref,breaklinks,colorlinks,allcolors=cvprblue]{hyperref}

\newcommand{\revised}[1]{\textcolor{black}{{#1}}}

\def\paperID{38205} 
\def\confName{CVPR}
\def\confYear{2026}

\title{Switch-JustDance: Benchmarking Whole-Body Motion Tracking \revised{Controllers} \\ Using a Commercial Console Game}

\author{Jeonghwan Kim$^{1*}$, Wontaek Kim$^{1*}$, Yidan Lu$^{2,7*}$, Jin Cheng$^{3*}$, Fatemeh Zargarbashi$^{3*}$, \\
Zicheng Zeng$^{4*}$, Zekun Qi$^{5*}$, Zhiyang Dou$^{6}$, Nitish Sontakke$^{1}$, Donghoon Baek$^{1}$,  \\
Sehoon Ha$^{1}$, Tianyu Li$^{1}$\\
$^{1}$Georgia Tech, $^{2}$HKU, $^{3}$ETH Zurich, $^{4}$SCUT, $^{5}$THU, $^{6}$MIT, $^{7}$PNDbotics \\
{\small *Equally Contributed, $\{jkim3662, wkim345, sehoonha, tli471\}@gatech.edu$}
}

\begin{document}

\makeatletter
\let\@oldmaketitle\@maketitle%
\renewcommand{\@maketitle}{\@oldmaketitle%
    \centering

    \centering
    \vspace{-2em}
        \includegraphics[width=0.95\textwidth]{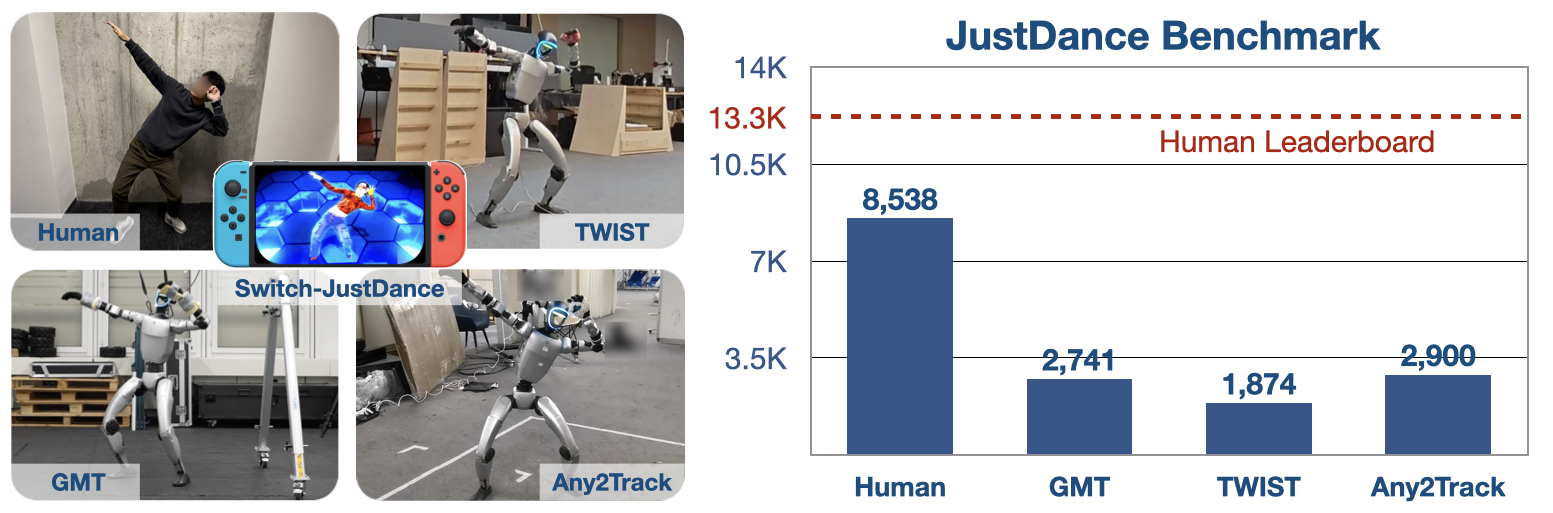}
        \vspace{-0.5em}
        \captionof{figure}{We introduce \textbf{Switch-JustDance}, a benchmark system for evaluating humanoid control policies using the \emph{Just Dance} game on the Nintendo Switch. Using this system, we benchmark three general humanoid controllers: GMT, TWIST and Any2Track, and compare their performance against human players.}
    \label{fig:front_page}
}
\makeatother

\maketitle
{
  \renewcommand{\thefootnote}{}
  \footnotetext{\revised{Project page: \url{https://switch-justdance.github.io/}}}
  \let\thefootnote\svthefootnote
}

\begin{abstract}

Recent advances in whole-body robot control have enabled humanoid and legged robots to perform increasingly agile and coordinated motions. However, standardized benchmarks for evaluating these capabilities in real-world settings, and in direct comparison to humans, remain scarce. Existing evaluations often rely on pre-collected human motion datasets or simulation-based experiments, which limit reproducibility, overlook hardware factors, and hinder fair human–robot comparisons. We present \textbf{Switch-JustDance}, a low-cost and reproducible benchmarking pipeline that leverages motion-sensing console games, \textit{Just Dance} on the Nintendo Switch, to evaluate robot whole-body control. Using \textit{Just Dance} on the Nintendo Switch as a representative platform, Switch-JustDance converts in-game choreography into robot-executable motions through streaming, motion reconstruction, and motion retargeting modules and enables users to evaluate controller performance through the game’s built-in scoring system. We first validate the evaluation properties of \textit{Just Dance}, analyzing its reliability, validity, sensitivity, and potential sources of bias. Our results show that the platform provides consistent and interpretable performance measures, making it a suitable tool for benchmarking embodied AI. Building on this foundation, we benchmark three state-of-the-art humanoid whole-body controllers on hardware and provide insights into their relative strengths and limitations. 
\end{abstract}

\vspace{-0.5cm}

\section{Introduction}
\label{sec:intro}

Recent progress in whole-body robot control has enabled humanoid and legged robots to demonstrate increasingly agile, coordinated, and human-like movements. Advances in both model-based whole-body controllers~\citep{khazoom2024tailoring,li2024cafe} and reinforcement learning~\citep{xie2025kungfubot,chen2025gmt,li2025amo,ze2025twist} have led to robots executing skills such as gymnastic motions, dynamic dancing, and real-time teleoperation. These developments mark a significant step toward embodied agents that can interact robustly with the world in athletic and versatile ways. As these capabilities evolve, a new and fundamental challenge is emerging for the community: how to evaluate the athletic performance of robot control policies in a fair, reproducible, and physically grounded way that allows direct comparison with humans.

Many recent works rely on large-scale human motion datasets such as AMASS~\citep{AMASS:ICCV:2019} and LaFAN1~\citep{harvey2020robust} to train motion imitation policies and evaluate whole-body controllers for humanoid and legged robots~\citep{he2024hover,Sferrazza2024HumanoidBench,Tessler2024MaskedMimic,shrestha2024generating}. 
While these datasets have been critical for enabling diverse motion learning, they are infrequently updated, and the choice of splits and evaluation protocols varies across studies, making consistent cross-method comparison difficult. 
Simulation benchmarks such as HumanoidBench~\citep{Sferrazza2024HumanoidBench} begin to standardize tasks, but remain restricted to specific robot models and simulated environmen  ts. 
Real-world benchmarking is even more challenging: motion-capture-based evaluation requires dedicated equipment, controlled environments, and high operational cost, and therefore large-scale or community-wide adoption remains impractical. Finally, robotic performance is rarely compared directly against humans under the same evaluation criteria, leaving open the question of how close robots are to approaching human-level athletic skill. These issues highlight the need for a standardized, reproducible, and accessible framework for benchmarking whole-body control in real-world conditions.

We present \textbf{Switch-JustDance}, a benchmarking framework that leverages the commercial motion-sensing game \emph{Just Dance} on the Nintendo Switch~\citep{nintendoSwitch,justdance} to evaluate whole-body humanoid controllers. The Nintendo Switch provides a low-cost ($\sim\$400$), plug-and-play setup that is robust for repeated use, while \emph{Just Dance} offers a large and continually updated library of full-body motions and a built-in scoring system that naturally accommodates players of different body sizes. Because the platform is designed for human users, it enables direct and quantitative comparisons between robots and humans under identical conditions, making it a promising candidate for standardized evaluation. In this work, we take a first step toward establishing its benchmarking potential by conducting a systematic validation study of the \emph{Just Dance} scoring system. We analyze its validity, discriminability, and repeatability through an extensive user study. Our results demonstrate that the scoring mechanism, which relies on the IMU-based Joy-Con controller, produces stable and interpretable performance measures, supporting its use as a reliable benchmark for embodied AI.

Building on this foundation, we instantiate Switch-JustDance on a humanoid platform and benchmark three representative state-of-the-art controllers: GMT~\citep{chen2025gmt}, 
TWIST~\citep{ze2025twist}, and Any2Track~\citep{zhang2025any2track}. By evaluating the performance of these controllers and comparing them against human performance, we show that current humanoid controllers still fall far short of human-level ability, and that Switch-JustDance reveals several important performance characteristics: smoothed input motions improve tracking stability, dynamic motions expose the limits of high-frequency whole-body control, and simulation results tend to overestimate controller robustness compared to hardware. These findings highlight both the challenges of dynamic whole-body dancing and the value of Switch-JustDance as a fair and physically grounded benchmarking tool.

In summary, this work makes three main contributions. First, we propose \textbf{Switch-JustDance}, a low-cost and accessible framework that uses the \emph{Just Dance} game on the Nintendo Switch as a standardized, real-world benchmark for evaluating whole-body humanoid controllers. Second, we perform a systematic validation of the platform’s scoring mechanism, demonstrating that it provides valid, discriminative, and repeatable performance measurements suitable for embodied AI evaluation. Third, we apply Switch-JustDance to benchmark three state-of-the-art humanoid controllers and provide insights into their capabilities and limitations when compared against human performance. Together, these contributions establish Switch-JustDance as a practical and physically grounded benchmark for advancing athletic whole-body robot control.
\vspace{-0.3cm}
\begin{figure*}
    \centering
    \includegraphics[width=0.95\linewidth]{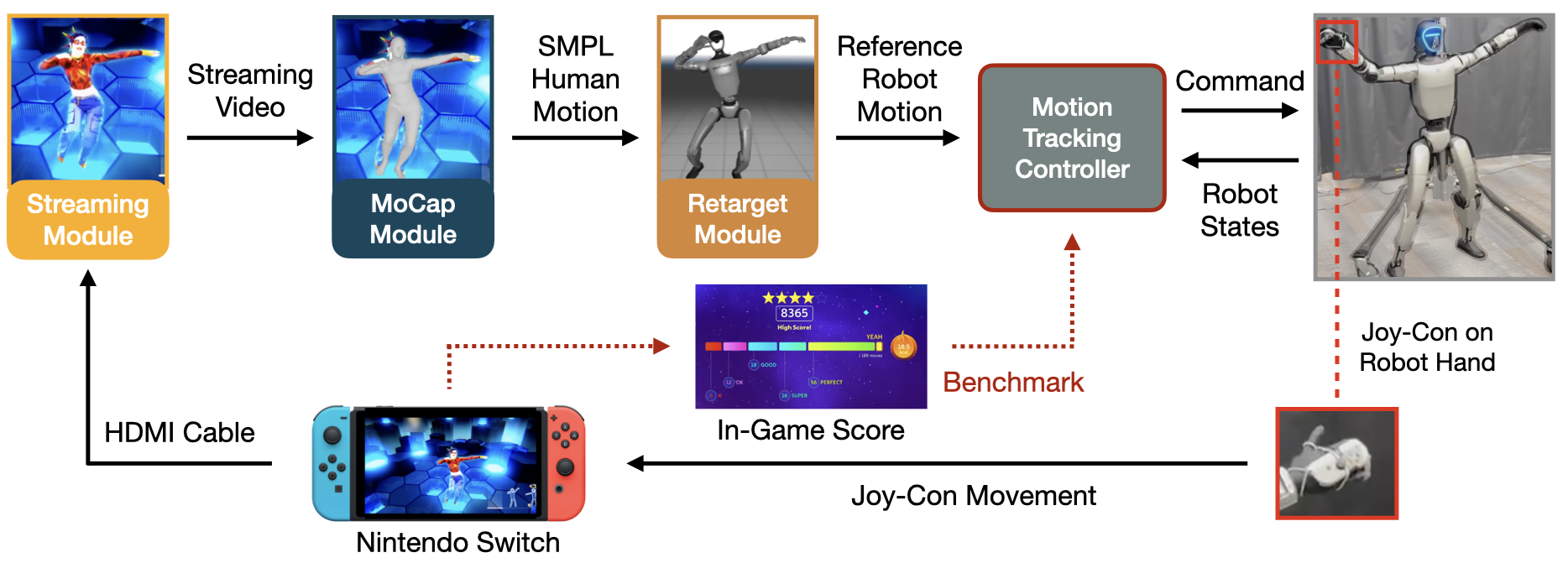}
    \caption{\small{\textbf{Switch-JustDance} captures Nintendo Switch gameplay and streams it to a MoCap module that recovers the dancer’s motion in SMPL human motion. The pose is retargeted to the robot via the retarget module, executed by a whole-body controller, and the robot’s performance is scored in-game.}
    }
    \label{fig:system_overview}
    \vspace{-0.5cm}
\end{figure*}

\section{Related Works}
\label{sec:related_works}

\subsection{Humanoid whole-body controllers}
Recent years have seen rapid progress in learning-based and optimization-based controllers for humanoid whole-body control. Learning approaches include versatile generalist policies that unify multiple control modes via motion-imitation distillation~\citep{he2024hover}, teleoperation-to-policy pipelines that track human motion on hardware~\citep{ze2025twist}, unified motion-tracking frameworks trained from large mocap datasets~\citep{chen2025gmt}, hybrid RL+trajectory-optimization methods for hyper-dexterous movements~\citep{li2025amo}, and controllers targeting highly dynamic skills such as kung-fu and dancing~\citep{xie2025kungfubot}. In parallel, optimization-centric systems push whole-body model predictive control toward real-time deployment, including cascaded-fidelity MPC with tuning-free whole-body control~\citep{li2024cafe} and accuracy-tailored fast whole-body NMPC for legged/humanoid robots~\citep{khazoom2024tailoring}.

Evaluation in these works typically combines (i) simulation metrics—pose/joint/EE tracking errors, stability and energy proxies—using mocap or keyframe references; and (ii) hardware demonstrations scored via task success or qualitative videos, sometimes with additional tracking metrics from external motion capture~\citep{he2024hover,chen2025gmt,li2025amo,xie2025kungfubot}. While these practices validate feasibility and absolute tracking quality, they are difficult to compare across papers due to heterogeneous datasets, reward/metric definitions, and task suites. Moreover, real-world evaluations that rely on optical MoCap demand dedicated spaces and instrumentation, which are costly and cumbersome to replicate at scale. As a result, many studies remain partly simulation-bound and rarely provide a human-vs-robot comparison under identical scoring—leaving open the question of how close current controllers are to human athletic performance.

\subsection{Commercial games for AI benchmarking}
Commercial games have long catalyzed progress in AI benchmarking and dataset creation. The Arcade Learning Environment (ALE) enabled reproducible, multi-task evaluation across dozens of Atari 2600 games~\citep{bellemare2013arcade}, underpinning breakthroughs such as DQN’s human-level control~\citep{mnih2015human}. Subsequent platforms broadened task complexity and partial observability (e.g., VizDoom~\citep{kempka2016vizdoom} and StarCraft II~\citep{vinyals2017starcraft}). Minecraft-based ecosystems advance open-ended embodied agents and sample-efficient RL through competitions and large knowledge bases (MineRL and MineDojo)~\citep{milani2020retrospective,kanervisto2022minerl,fan2022minedojo}. For driving perception, Grand Theft Auto V has been exploited to generate large, pixel-accurate synthetic datasets and labels at scale, accelerating training and reducing annotation cost~\citep{richter2016playing}.  Across these examples, games provide standardized interfaces, abundant content, and community baselines. Our work follows this lineage but targets physically grounded, hardware-in-the-loop benchmarking for embodied AI.

\subsection{VR and console platforms for embodied AI}
Commodity gaming hardware has long been used to lower barriers in embodied AI research. Microsoft Kinect enabled low-cost full-body motion capture and was widely adopted for human–robot interaction, gesture-based teleoperation, and benchmarking tasks~\citep{shotton2011real,sturm2012benchmark}. Similarly, Nintendo Wii devices, including Wii-motes and the Balance Board, were employed in rehabilitation and HRI studies to capture motion and balance signals for interactive training~\citep{burke2009optimising}. These works demonstrated that consumer-level consoles can meaningfully support motion capture and embodied evaluation, albeit often with domain-specific or limited reproducibility. More recently, VR-based platforms have emerged as flexible alternatives to optical MoCap for humanoid learning~\citep{winkler2022questsim, ahuja2022controllerpose}. Systems such as Open-TeleVision~\citep{cheng2024open}, H2O~\citep{he2024learning}, OmniH2O~\citep{he2024omnih2o}, and OpenWBT~\cite{zhang2025openwbt} enable immersive whole-body teleoperation and large-scale demonstration collection for humanoid robots using head-mounted displays and consumer sensors. These pipelines provide naturalistic embodiment and scalable data collection, but still require careful calibration and custom infrastructure. Our work builds on this trajectory by exploring motion-sensing console games with built-in scoring. In particular, we investigate \emph{Just Dance} on the Nintendo Switch, which uniquely combines continuous motion updates, standardized quantitative evaluation, and direct human–robot comparability—positioning it as a promising benchmark for whole-body robotic control.

\section{Switch-JustDance}
\label{sec:method}

We introduce \textbf{Switch-JustDance}, a benchmark framework that enables general humanoid motion controllers to play the \emph{Just Dance} game on the Nintendo Switch platform, thereby using the in-game score as an objective measure of controller performance. \emph{Just Dance} is a rhythm-based game in which players imitate on-screen choreography and receive real-time scores based on their motion similarity to the in-game demonstrator. A humanoid controller that closely reproduces the demonstrator’s movement achieves a higher score, while inaccurate or unstable control results in a lower score.

An overview of the pipeline is shown in Figure~\ref{fig:system_overview}. The system consists of four main modules:

\begin{itemize}
    \item \textbf{Streaming Module:} Captures the Nintendo Switch’s real-time gameplay via an HDMI capture card and streams the video frames to the motion capture system once the game starts with negligible latency.
    
    \item \textbf{MoCap Module:} Reconstructs 3D human poses from monocular RGB video using GVHMR~\citep{GVHMR}, which outputs SMPL~\citep{SMPL:2015} parameterizations of the dancer’s motion. We modify GVHMR for real-time operation using a sliding-window inference scheme, achieving stable pose estimates at approximately 200~ms per frame.
    
    \item \textbf{Retargeting Module:} Maps the reconstructed SMPL motions to the target robot morphology using GMR~\citep{ze2025gmr}, ensuring kinematic feasibility across robots of varying proportions. GMR runs at roughly 50~ms per frame and produces robot-space keyframe trajectories synchronized with the MoCap updates. By using an asynchronous interpolation thread, intermediate frames are interpolated using linear interpolation (LERP) for positions and spherical interpolation (SLERP) for rotations, increasing the output frequency between GVHMR/GMR frames.
    
    \item \textbf{Motion Tracking \revised{Controller}:} Generates control commands based on the retargeted reference motion and the robot’s proprioceptive feedback. This \revised{control policy} represents the humanoid controller under evaluation.
\end{itemize}

A Joy-Con controller is securely mounted to the robot’s hand. As the robot moves, the Joy-Con sends motion signals to the Nintendo Switch, allowing it to track the robot’s movements and compute in-game scores identically to a human participant.

We implement and test the full pipeline on a workstation equipped with an RTX~4090 GPU and an Intel Core~i9-13900K CPU. The end-to-end runtime is dominated by GVHMR inference ($\sim$200~ms per step), with GMR retargeting adding about 50~ms per keyframe. The asynchronous interpolation thread converts these sparse updates into a smooth 10--15~Hz reference stream, ensuring that the evaluated controllers receive temporally consistent targets with low additional overhead. The details of the asynchronous implementation are presented in the Appendix~\ref{sec:asynch_system}.


Compared with existing evaluation systems, Switch-JustDance offers several advantages. First, it is low cost: the entire setup requires only a Nintendo Switch (about $\$$400), while even entry-level motion-capture systems typically cost over $\$$50,000. Second, it is easy to set up and highly repeatable, as the Switch platform is globally available and adheres to a consistent standard across all regions. Third, the system naturally supports cross-embodiment evaluation. Because \textit{Just Dance} is designed for players of different body sizes—including children, adults, and now robots—it does not require motion retargeting to compute scores. Finally, the game provides long-horizon (over 3 minutes), dynamic, and high-quality dance motions, with new content added every year. In contrast, widely used motion datasets such as AMASS~\citep{AMASS:ICCV:2019} are relatively static.

Before using this system as a benchmarking platform, an important question arises: does the built-in scoring of \emph{Just Dance} provide consistent and meaningful feedback that correlates with controller quality across both humans and robots? The following section investigates this question through systematic benchmarking and metric analysis.

\section{System Validation}
\label{sec:system_analysis}

We conducted a systematic validation study to evaluate the properties of Switch-JustDance's. Specifically, we aimed to address the following three questions: (1) \textbf{Validity:} Does the pipeline’s metric—the \textit{Just Dance} score—serve as a valid indicator of humanoid controller performance? (2) \textbf{Discriminability:} Is the score sufficiently discriminative to differentiate players with different skill levels? and (3) \textbf{Repeatability:} Is the pipeline stable enough to yield consistent results across repeated trials? In addition, we examined potential limitations of Switch-JustDance.

The validation was performed through an extensive user study involving ten human participants (ages 20–34) with diverse genders and body types. Each participant performed five songs of varying difficulty levels, repeating each song four times. This setup resulted in 20 dance trials per participant and a total of 200 recorded dance motions for analysis.

\vspace{-0.4cm}

\paragraph{Validity}
To assess the validity of the \textit{Just Dance} Score (JDS) as a quantitative metric for humanoid motion tracking, we compare the collected JDS values against a standard kinematic-based metric commonly used in motion tracking evaluations:  Procrustes-Aligned Mean Per-Joint Position Error (PA-MPJPE)\cite{Kendall1989ShapeSurvey}. PA-MPJPE measures the average joint position error between the player and a reference motion after applying Procrustes transformation (rigid and scale) in the player's coordinate. This metric is computed by aligning each participant’s recorded motion with the reference choreography reconstructed by the MoCap Module and then calculating the mean joint position error to quantify tracking accuracy.

Our analysis shows that the correlation coefficient ($r$) and significance level ($p$) between JDS and PA-MPJPE across different songs range from $r = [-0.76, -0.42]$ and $p = [<0.001, <0.05]$, clearly indicating that higher JDS values correspond to lower tracking errors, validating JDS as a reliable indicator of motion-tracking quality. 
Interestingly, the results also imply that JDS captures additional aspects of performance beyond pure kinematic accuracy. According to the related patent document~\cite{us9358456b1}, the JDS computation also considers features such as joint acceleration and rhythmic alignment with the music beat—important perceptual and dynamic qualities not reflected in PA-MPJPE alone.

\vspace{-0.3cm}
\paragraph{Discriminability}
An effective evaluation metric should clearly distinguish between high- and low-quality performances. To assess the discriminative capability of the JDS, we compared three reference levels: (1) a static baseline in which the Joy-Con controllers remained stationary, representing the lower bound; (2) the world leaderboard scores, representing the upper bound achieved by expert human players; and (3) the averaged scores of our study participants. The static baseline consistently yielded scores near 0, while the global leaderboard scores reached a maximum score of 13{,}333. In contrast, our participants achieved intermediate scores ranging between [3460, 12{,}952]. This wide separation between the lower, middle, and upper ranges demonstrates that JDS provides sufficient dynamic range to distinguish motion performances of varying quality, supporting its utility as a discriminative evaluation metric for benchmarking humanoid controllers.

\vspace{-0.3cm}

\paragraph{Repeatability}
To assess the repeatability of the JDS, we examined how consistently it scores repeated executions of the same dance motion. We computed the Intraclass Correlation Coefficient (ICC)\cite{shrout1979intraclass} and the Coefficient of Variation (CV)\cite{fisher1970statistical}. A higher ICC (closer to 1.0) and a lower CV indicate stronger consistency in repeated measurements. Our analysis shows that the JDS achieved an ICC of 0.735 and a CV of 10.0\%, indicating low trial-to-trial variation and demonstrating that the metric provides reliable, repeatable evaluations.

In addition to ICC and CV, we computed Kendall’s Coefficient of Concordance (KCC)\cite{kendall1939problem} across multiple songs and repetitions to assess ranking consistency among participants. KCC quantifies the agreement in rank ordering across different conditions, where $KCC = 1$ indicates perfect agreement and $KCC = 0$ represents random ranking. The obtained $KCC = 0.678$ demonstrates that participants who ranked higher in one dance tended to maintain similar rankings across other dances and repetitions. Together, these results confirm the strong repeatability and robustness of the Switch-JustDance benchmarking pipeline.

\vspace{-0.3cm}

\paragraph{Potential Shortcoming}
One limitation of the current benchmarking pipeline is that the \textit{Just Dance} scoring system relies primarily on the motion of a single Joy-Con, allowing players to achieve high scores through upper-body motion while keeping the lower body static. 
\revised{However, achieving high scores implicitly requires coherent full-body coordination, as rhythm, balance, and timing in dance are strongly coupled under the assumption that the control policy is tasked to faithfully track the reference motion.}
In future scenarios, we encourage participants to provide videos to ensure visual fidelity.

\begin{figure*}
    \centering
    \includegraphics[width=0.95\linewidth]{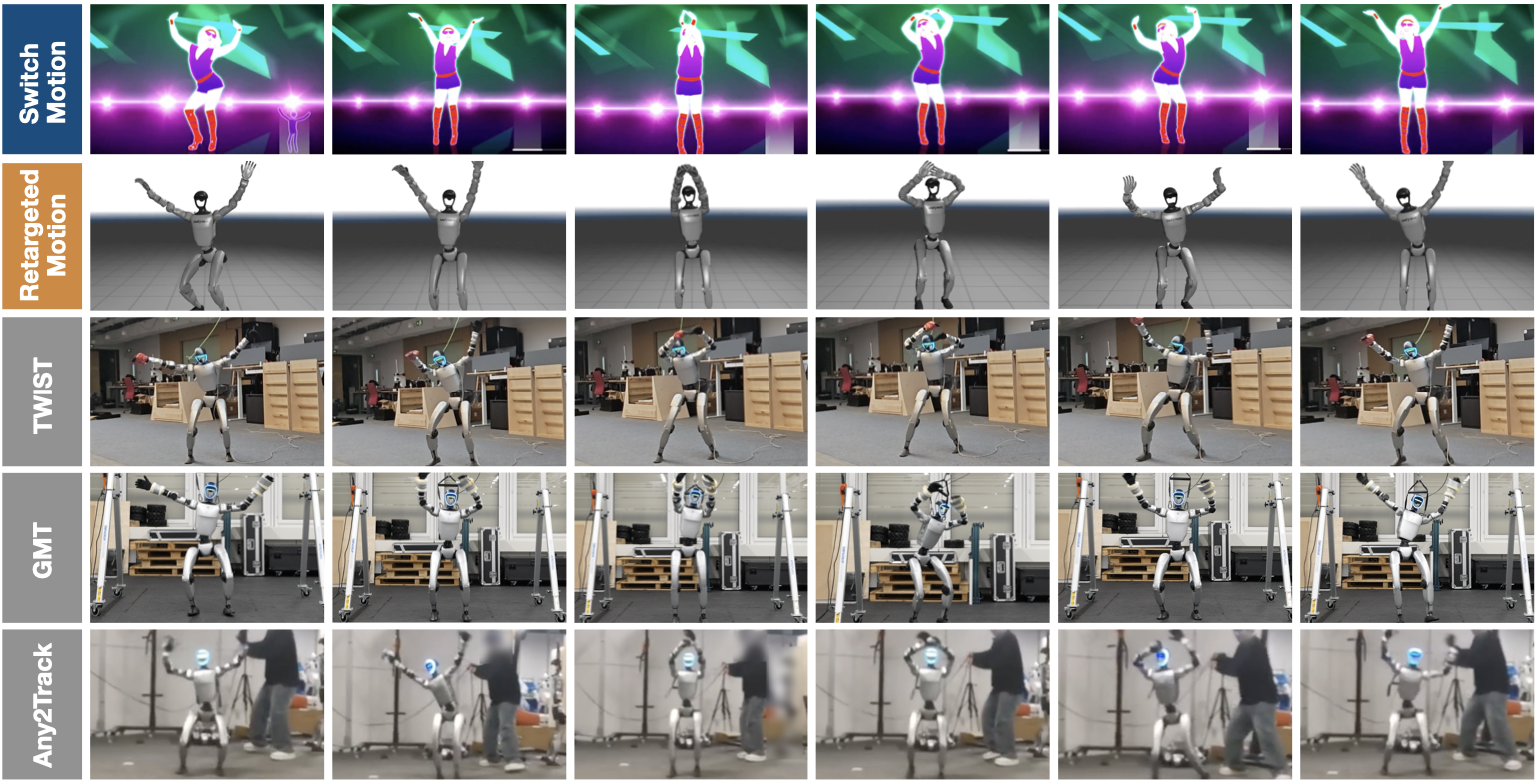}
    \caption{\small{Top to bottom: Switch motion as source frames, retargeted motion from GMR output, and three humanoid controllers on hardware (TWIST, GMT, and Any2Track). Columns progress left to right in time. Unitree G1 plays \emph{Just Dance} by holding a Joy-Con, enabling in-game scoring.}
    }
    \label{fig:snapshots}
    \vspace{-0.5cm}
\end{figure*}

\section{Benchmark Metrics and Experiments}
\label{sec:Benchmark}

Having established the reliability of Switch-JustDance as an evaluation platform, we now employ it to benchmark humanoid whole-body control policies. This section first introduces the evaluation metrics, then describes the experimental setups, and finally outlines the candidate motion controllers used in our study.

\subsection{Benchmark Metrics}
\label{sec:benchmark_metrics}
We use four complementary metrics to evaluate general humanoid motion-tracking controllers, each capturing a distinct aspect of performance:

\begin{itemize}
    \item \textbf{Just Dance Score~(JDS):} The in-game score from the Nintendo Switch serves as our primary performance indicator. Computed from Joy-Con motion tracking, it reflects how well the robot synchronizes with the reference choreography. The max score for every song in \textit{Just Dance} is 13{,}333.
    
    \item \textbf{Mean Per Joint Position Error~(MPJPE, mm):} We measure the average joint position error between the robot’s state estimation and the retargeted reference trajectory in the robot's torso frame. Lower values indicate more accurate and consistent pose tracking. Since the errors are measured in the torso frame, we omit Procrustes alignment. Because the robot may fall mid-dance, we report MPJPE in two ways: \emph{Active} is computed only over frames before a fall, and \emph{All} is computed over the entire sequence, including any post-fall frames. If no fall occurs, the two are identical.
    
    \item \textbf{Success Rate~(SR, \%):} We record the percentage of trials in which the controller completes the entire dance without falling or overheating. This metric highlights the controller’s stability and robustness over long-horizon motions.
    
    \item \textbf{Smoothness:} Smoothness is quantified by the average joint jerk(rad/s$^3$)~\cite{jerk} and acceleration(rad/s$^2$) of the joint angles computed using finite difference. Controllers that produce jittery or impulsive motion yield higher jerk, even when MPJPE is low. This metric emphasizes natural, human-like movement and hardware-friendly actuation.
\end{itemize}

\subsection{Experimental Settings}

We evaluate controllers across several experimental settings, each reflecting a different level of input fidelity:

\begin{enumerate}
    
    \item \textbf{Streaming Smoothed Gameplay~(Smo):} This setting imitates the online gameplay of \emph{Just Dance}. The MoCap Module captures human motion in real time and retargets it on the fly. However, real-time streaming provides no future reference poses and offers only a small computation window, which can cause jitter in the reconstructed poses. To address this, we apply a temporal filter to the pose stream using low-pass smoothing with outlier rejection and velocity clamps. This produces a conservative but smooth reference motion, while reducing high-frequency details.

    \item \textbf{Offline Dynamic Reference~(Dyn):} 
    This setting uses refined, preprocessed trajectories that restore the high-frequency motions removed in the \textit{Smo} setting. The trajectories are first captured from the unfiltered streaming motion. We then process the full sequence offline to improve temporal smoothness, kinematic continuity, and coordinate-frame consistency. Specifically, we run GVHMR in offline batch mode, resample the motion to 30~Hz using SLERP/LERP, and correct all frames before applying GMR retargeting. Unlike the online setting, which requires heavy causal filtering for low-latency operation, \textit{Dyn} uses non-causal smoothing that preserves timing accuracy and sharp motion transients. As a result, it provides a cleaner but more dynamically challenging reference. During evaluation, we manually align the game start time with the controller input stream.
\end{enumerate}

\begin{table*}[t]
\centering
\setlength{\tabcolsep}{6pt}
\resizebox{0.9\linewidth}{!}
{
\begin{tabular}{lcccccccccccc}
\toprule
\textbf{Player-Setting} & \multicolumn{3}{c}{\textbf{JDS} $\uparrow$} & \multicolumn{2}{c}{\textbf{MPJPE(mm)} $\downarrow$} & \multicolumn{2}{c}{\revised{\textbf{DTW(mm)} $\downarrow$}} & \textbf{SR(\%)} $\uparrow$ & \multicolumn{2}{c}{\textbf{Smoothness} $\downarrow$} \\
\textbf{} & Easy & Hard & All & Active & All & \revised{Act} & \revised{All} & & Jerk(rad/s$^3$) & Acc(rad/s$^2$) \\
\midrule
Human-leaderBoard & 13333 & 13333 & 13333 & - & - & - & - & 100.0 & - & - \\
Human & 9867 & \revised{6543} & \revised{8538} & \revised{111.9} & \revised{111.9} & \revised{102.4} & \revised{102.4} & 100.0 & \revised{3350.5} & \revised{95.1} \\
\midrule
GMT-Real-Smo & 2765 & 1891 & 2415 & 130.9 & 139.5 & \revised{114.1} & \revised{120.3} & 73.3 & 3174.1 & 78.5 \\
TWIST-Real-Smo & 2085 & 1557 & 1874 & 117.1 & 119.8 & \revised{96.9} & \revised{97.9} & 80.0 & 2811.3 & 70.6 \\
Any2Track-Real-Smo & 2978 & 2785 & 2900 & 102.6 & 139.4 & \revised{44.7} & \revised{69.1} & 60.0 & 1978.8 & 82.6 \\
\midrule
GMT-Real-Dyn & 2881 & 2530 & 2741 & 88.8 & 102.6 & \revised{50.8} & \revised{59.7} & 80.0 & 4098.6 & 126.3 \\
TWIST-Real-Dyn & 1501 & 1140 & 1357 & 108.5 & 142.7 & \revised{71.0} & \revised{95.4} & 33.3 & 4024.1 & 126.0 \\
Any2Track-Real-Dyn & 2753 & 2251 & 2552 & 114.2 & 128.2 & \revised{40.9} & \revised{49.5} & 80.0 & 3783.2 & 127.6 \\
\midrule
GMT-Sim-Smo & - & - & - & \revised{121.0} & 137.4 & \revised{105.9} & \revised{118.5} & 20.0 & 2146.2 & 58.7 \\
TWIST-Sim-Smo & - & - & - & 116.7 & 125.2 & \revised{106.5} & \revised{108.4} & 80.0 & 1684.6 & 49.4 \\
Any2Track-Sim-Smo & - & - & - & 91.4 & 106.9 & \revised{65.2} & \revised{69.2} & 80.0 & 2115.1 & 88.1 \\
\midrule
GMT-Sim-Dyn & - & - & - & 87.8 & 137.0 & \revised{74.7} & \revised{114.0} & 0.0 & 3694.6 & 110.5 \\
TWIST-Sim-Dyn & - & - & - & 100.9 & 127.8 & \revised{62.2} & \revised{74.7} & 40.0 & 3438.9 & 110.1 \\
Any2Track-Sim-Dyn & - & - & - & 90.9 & 105.7 & \revised{65.6} & \revised{65.5} & 80.0 & 3949.8 & 130.1 \\
\bottomrule
\end{tabular}
}
\caption{\textbf{Benchmark results across controllers, input types, and simulation settings.} 
We evaluate three representative humanoid controllers (GMT, TWIST, and Any2Track) under smoothed (\textit{Smo}) and dynamic (\textit{Dyn}) input motions, as well as their simulation counterparts. Higher \textit{Just Dance} Score (JDS) indicates better task performance, \revised{lower MPJPE and Dynamic Time Warping (DTW) indicate more accurate motion tracking and temporal alignment,} and higher Success Rate (SR) reflects improved stability. Smoothness is measured using acceleration and jerk. 
}
\vspace{-0.5cm}
\label{tab:benchmark_result}
\end{table*}

In summary, \textit{Smo} favors stability by suppressing high-frequency content at the cost of a small phase lag, 
whereas \textit{Dyn} preserves timing and dynamics, which can improve choreography matching but challenges controller agility. 

We evaluate both \textit{Smo} and \textit{Dyn} settings in both simulated (Sim) and real-world (Real) environments to evaluate the Sim to Real performance gap. Because no physical Joy-Con is present, the \textit{Just Dance} Score is not applicable in simulation. Each controller is evaluated on five distinct dance routines described in Sec.~\ref{sec:system_analysis}, comprising three ``easy'' (Level~1--2) and two ``hard'' (Level~3--4) dances. For each routine, we perform three repeated trials and record both the \emph{Just Dance Score} and full-body joint trajectories obtained by internal state estimator. In total, each controller is evaluated over \(60\) trials \((4 \times 5 \times 3)\). During experiments, gantry is used for the safty of the robot.

\subsection{Candidate Controllers}

We benchmark three state-of-the-art general-purpose whole-body tracking humanoid controllers, each trained on distinct datasets and architectures. To the best of our knowledge, none of the controllers leverages motion data from the \textit{Just Dance} game.
We use checkpoints provided by the respective authors for evaluation.

\begin{enumerate}
    \item \textbf{GMT~\citep{chen2025gmt}:} learns a unified motion-tracking policy by leveraging adaptive sampling and a mixture-of-experts architecture for diverse whole-body motions. For \emph{Smo}, we do not expose future frames and use a constant hold-last-frame assumption over the lookahead. For \emph{Dyn}, future reference frames are available and included in the observation stack.
    
    \item \textbf{TWIST~\citep{ze2025twist}:}  leverages human motion capture to train a general robot motion controller for teleoperation tasks.
    
    \item \textbf{Any2Track~\citep{zhang2025any2track}:} learns a tracking policy using a two-stage RL framework that augments a general tracker with online disturbance adaptation to handle varied real-world conditions.
\end{enumerate}

We additionally report human baseline alongside the benchmarked humanoid controllers.
The world leaderboard entries of JDS are obtained directly from the official in-game leaderboard.
We also compute the average JDS and kinematic metrics from the 200 dance motions collected in our user study.
For kinematic metrics, human motion sequences are retargeted to the robot’s joint space to eliminate biases arising from differing joint configurations.

\section{Benchmark Results}
\label{sec:results}

Overall, our proposed Switch-JustDance system enables humanoid controllers to execute the dance motions shown in the game, and the \textit{Just Dance} platform provides a real-time score for evaluating their performance. 
Figure~\ref{fig:snapshots} and in the supplementary materials, and Table~\ref{tab:benchmark_result} summarize the quantitative benchmark results.

The results indicate that current humanoid controllers are incomparable to human-level performance. A human expert can reach the theoretical maximum score of the game, and participants in our user study achieve an average score of \revised{8538}. In comparison, even the best-performing controller reaches only about one-third of the human baseline score (Any2Track 2900 vs.\ human \revised{8538}). Interestingly, when comparing human motion against all controllers in the \textit{Real-Dyn} setting, we find that these controllers actually achieve lower MPJPE than human players, with the best MPJPE reaching 88.8~mm while human players get \revised{111.9}~mm. This suggests that the robots can track the reference pose more closely at the frame level than human participants. However, this improved accuracy comes at the cost of stability: robots with lower MPJPE also exhibit lower success rates. Furthermore, human players produce much smoother movements, showing lower jerk and acceleration compared to all controllers.

Examining the performance of the three controllers, we find that GMT achieves relatively high \textit{Just Dance} Scores and success rates in  \textit{Real-Dyn} settings. TWIST achieves strong performance in terms of MPJPE, success rate, and smoothness under the \textit{Real-Smo} setting, but does not match GMT in terms of JDS. This highlights that JDS is a more sophisticated metric that does not simply depend on pose tracking accuracy or motion smoothness. Any2Track gets the highest overall JDS in the \textit{Real-Smo} setting, it also show smoother motions compare to GMT and TWIST. Any2Track is also the only controller that can finish entire dance without gantry.

We observe clear effects of song difficulty. All controllers perform better on the ``Easy'' routines, achieving higher JDS and more stable tracking. On the ``Hard'' routines, the JDS drops significantly for every controller, with an average decrease of 573 points. This shows that higher-dynamical motion changes and more complex coordination patterns pose substantial challenges for current whole-body control policies.

Overall, the benchmark highlights a clear performance gap between robots and humans and reveals meaningful differences between existing humanoid controllers. These results demonstrate that the proposed benchmark provides a sensitive and physically grounded way to evaluate whole-body control performance in dynamic, real-world scenarios.

\subsection{Effect of Input Motion~(Real-Smo vs.\ Real-Dyn)}

Our experiments involve two input types. In the \textit{Real-Smo} setting, the input motions are smoothed to remove high-frequency components, which reduces rapid changes and emphasizes stability. In contrast, the \textit{Real-Dyn} experiments use input motions that contain the original dynamic movements from the game, including fast swings and sudden posture changes.
We compare the results between the \textit{Real-Smo} and \textit{Real-Dyn} settings to study how the characteristics of the input motions affect the performance of the control policies. 

Across all controllers, we observe clear differences in the smoothness metrics. The \textit{Real-Dyn} inputs lead to significantly higher jerk and acceleration, since the robot must reproduce the rapid and abrupt transitions present in the raw dance motions. In contrast, \textit{Real-Smo} inputs result in gentler and more predictable motions, leading to smoother trajectories. On average, smoothing reduces jerk by 1268.7 compared to the \textit{Real-Dyn} setting.

High dynamic movements also increase the likelihood of instability and falling. For example, TWIST shows a large performance drop in the \textit{Real-Dyn} setting: its success rate decreases by 46.7\%, MPJPE increases by 23.9~mm, and JDS decreases by 16.7\%. 

In contrast to TWIST, Any2Track and GMT’s stability is less affected by dynamic input motions. GMT maintains a success rate above 70\% in both settings, showing that its control strategy is more robust to rapid movement changes. Furthermore, GMT achieves higher JDS in the \textit{Real-Dyn} setting than in \textit{Real-Smo}. This is because smoothing the motion introduces a mismatch between the executed motion and the original in-game choreography, which increases MPJPE and reduces score when the robot is actively tracking the input.

Overall, these observations show that input motion quality influences controller behavior. Smoothed inputs improve stability and therefore increase the overall dance performance, while dynamic inputs better reflect the challenges of real dance movements but can lead to instability of current whole-body controllers.

\subsection{Sim-to-Real Performance}
We study the sim-to-real performance of the controllers by comparing their hardware results with their corresponding simulation results. In general, simulated motions exhibit lower jerk and acceleration, while hardware executions show larger spikes. This difference is expected and is likely caused by real-world factors such as joint backlash, torque limits, friction, and timing delays in the low-level controller, all of which introduce additional noise and disturbances that do not exist in simulation.

The MPJPE difference between simulation and hardware is much smaller than the difference observed in smoothness metrics. This reveals a potential limitation of MPJPE as an evaluation metric. A controller may achieve high positional tracking accuracy (low MPJPE) while still producing shaky or unstable motions that compromise balance and could potentially damage the robot. 

Counterintuitively, we also observe lower success rates in simulation than on hardware. This is primarily because the physical robot receives slight support from the safety gantry, which prevents falls during critical moments, even though we aim to minimize its influence. The effect becomes more noticeable when global drift occurs, as the gantry can help the robot re-stabilize during execution. As a result, some trials that would fail in simulation are counted as successful on hardware. This behavior is observed more often in GMT, which is more sensitive to small disturbances, whereas TWIST and Any2Track show more consistent balancing and are less affected by the gantry.

\section{Discussion}
\label{sec:discussion}

In this work, we introduced \textbf{Switch-JustDance}, a low-cost and physically grounded benchmarking pipeline that uses commercial motion-sensing console games to evaluate the agility and accuracy of whole-body humanoid controllers. 
Using \textit{Just Dance} on the Nintendo Switch, our system enables direct comparison between robots and humans through in-game scores while integrating open-source motion capture and retargeting modules for real-time execution on a Unitree~G1. The benchmark results show a clear gap between current controllers and human performance, and highlight several key insights: smoothed motion inputs lead to more stable tracking, dynamic motions expose limitations in handling high-frequency movements, and simulation often overestimates stability compared to hardware. We also observe that the safety gantry can occasionally prevent falls, slightly inflating success rates on hardware. Overall, Switch-JustDance offers a sensitive and comprehensive evaluation framework by enabling fair human–robot comparisons, providing real-time assessment of whole-body tracking and balance, and supporting evaluation under fully OOD motions. We believe these features will contribute to advancing dynamic humanoid control and embodied AI.

We identify several promising directions for future research. First, current controllers, though capable of executing diverse motions, still lack smooth transitions and robustness under noisy or partially observed inputs for long and highly dynamic sequences. Advancing learning-based or hybrid control architectures to handle these challenges will be key to closing the human–robot performance gap.
In addition, enhancing motion reconstruction and retargeting algorithms represents another meaningful research direction. The current Switch-JustDance pipeline relies on open-source motion capture and retargeting modules, which are designed for broader scenarios. Future work could focus on faster, low-latency motion tracking and on retargeting methods that ensure dynamically consistent and physically grounded motion generation across different morphologies.
Finally, we can expand the scope of console-based embodied AI benchmarks. Beyond motion imitation, commercial game platforms provide a rich and diverse set of physically interactive environments. Leveraging these games as evaluation or training grounds could open new avenues for embodied AI.


\section{Acknowledgement}
\label{sec:acknowledgement}
This work was supported by Samsung Research.

{
    \small
    \bibliographystyle{ieeenat_fullname}
    \bibliography{main}

@String(CVPR= {IEEE Conf. Comput. Vis. Pattern Recog.})

@String(ICCV= {Int. Conf. Comput. Vis.})

@String(ECCV= {Eur. Conf. Comput. Vis.})

@String(TOG= {ACM Trans. Graph.})

@String(VR   = {Vis. Res.})

@String(CVPR  = {CVPR})

@String(ICCV  = {ICCV})

@String(ECCV  = {ECCV})

@String(TOG   = {ACM TOG})

@article{khazoom2024tailoring,
  title={Tailoring solution accuracy for fast whole-body model predictive control of legged robots},
  author={Khazoom, Charles and Hong, Seungwoo and Chignoli, Matthew and Stanger-Jones, Elijah and Kim, Sangbae},
  journal={IEEE Robotics and Automation Letters (RA-L)},
  year={2024},
  publisher={IEEE}
}

@article{li2024cafe,
  title={Cafe-mpc: A cascaded-fidelity model predictive control framework with tuning-free whole-body control},
  author={Li, He and Wensing, Patrick M},
  journal={IEEE Transactions on Robotics (T-RO)},
  year={2024},
  publisher={IEEE}
}

@patent{us9358456b1,
  author       = {Chabert, Charles and Cario, David},
  title        = {Dance Competition Game},
  number       = {US9358456B1},
  year         = {2016},
  month        = jun,
  assignee     = {Ubisoft Entertainment SA},
  url          = {https://patents.google.com/patent/US9358456B1},
  note         = {U.S. Patent and Trademark Office}
}

@article{xie2025kungfubot,
  title={KungfuBot: Physics-Based Humanoid Whole-Body Control for Learning Highly-Dynamic Skills},
  author={Xie, Weiji and Han, Jinrui and Zheng, Jiakun and Li, Huanyu and Liu, Xinzhe and Shi, Jiyuan and Zhang, Weinan and Bai, Chenjia and Li, Xuelong},
  journal={Advances in Neural Information Processing Systems (NeurIPS)},
  year={2025}
}

@article{ze2025twist,
title={TWIST: Teleoperated Whole-Body Imitation System},
author= {Yanjie Ze and Zixuan Chen and João Pedro Araújo and Zi-ang Cao and Xue Bin Peng and Jiajun Wu and C. Karen Liu},
year= {2025},
journal= {Conference on Robot Learning (CoRL)}
}

@article{zhang2025any2track,
  title={Track Any Motions under Any Disturbances}, 
  author={Zhikai Zhang and Jun Guo and Chao Chen and Jilong Wang and Chenghuai Lin and Yunrui Lian and Han Xue and Zhenrong Wang and Maoqi Liu and Jiangran Lyu and Huaping Liu and He Wang and Li Yi},
  journal={arXiv:2509.13833},
  year={2025},
}

@article{chen2025gmt,
title={GMT: General Motion Tracking for Humanoid Whole-Body Control},
author={Chen, Zixuan and Ji, Mazeyu and Cheng, Xuxin and Peng, Xuanbin and Peng, Xue Bin and Wang, Xiaolong},
journal={arXiv:2506.14770},
year={2025}
}

@article{li2025amo,
title={AMO: Adaptive Motion Optimization for Hyper-Dexterous Humanoid Whole-Body Control},
author={Li, Jialong and Cheng, Xuxin and Huang, Tianshu and Yang, Shiqi and Qiu, Rizhao and Wang, Xiaolong},
journal={Robotics: Science and Systems (RSS)},
year={2025}
}

@article{he2024hover,
title={HOVER: Versatile Neural Whole-Body Controller for Humanoid Robots},
author={He, Tairan and Xiao, Wenli and Lin, Toru and Luo, Zhengyi and Xu, Zhenjia and Jiang, Zhenyu and Liu, Changliu and Shi, Guanya and Wang, Xiaolong and Fan, Linxi and Zhu, Yuke},
journal={IEEE International Conference on Robotics and Automation (ICRA)},
year={2024}
}

@software{ze2025gmr,
title={GMR: General Motion Retargeting},
author= {Yanjie Ze and João Pedro Araújo and Jiajun Wu and C. Karen Liu},
year= {2025},
url= {https://github.com/YanjieZe/GMR},
note= {GitHub repository}
}

@inproceedings{GVHMR,
  title={World-grounded human motion recovery via gravity-view coordinates},
  author={Shen, Zehong and Pi, Huaijin and Xia, Yan and Cen, Zhi and Peng, Sida and Hu, Zechen and Bao, Hujun and Hu, Ruizhen and Zhou, Xiaowei},
  booktitle={ACM Transactions on Graphics (The proceeding of SIGGRAPH Asia)},
  pages={1--11},
  year={2024}
}

@conference{AMASS:ICCV:2019,
  title = {{AMASS}: Archive of Motion Capture as Surface Shapes},
  author = {Mahmood, Naureen and Ghorbani, Nima and Troje, Nikolaus F. and Pons-Moll, Gerard and Black, Michael J.},
  booktitle = {IEEE International Conference on Computer Vision (ICCV)},
  pages = {5442--5451},
  month = oct,
  year = {2019},
  month_numeric = {10}
}

@article{harvey2020robust,
author    = {Félix G. Harvey and Mike Yurick and Derek Nowrouzezahrai and Christopher Pal},
title     = {Robust Motion In-Betweening},
journal = {ACM Transactions on Graphics (TOG)},
publisher = {ACM},
volume    = {39},
number    = {4},
year      = {2020}
}

@software{nintendoswitch,
  author = {{Nintendo}},
  title = {{Nintendo Switch}},
  note = {Video game console},
  year = {2017},
  url = {https://www.nintendo.com/switch/},
}

@software{justdance,
  author = {{Ubisoft}},
  title = {{Just Dance}},
  note = {Video game},
  year = {2009},
  url = {https://www.ubisoft.com/en-us/game/just-dance/},
}

@article{SMPL:2015,
      author = {Loper, Matthew and Mahmood, Naureen and Romero, Javier and Pons-Moll, Gerard and Black, Michael J.},
      title = {{SMPL}: A Skinned Multi-Person Linear Model},
      journal = {ACM Transactions on Graphics (The proceeding of SIGGRAPH Asia)},
      month = oct,
      number = {6},
      pages = {248:1--248:16},
      publisher = {ACM},
      volume = {34},
      year = {2015}
    }

@article{bellemare2013arcade,
  title={The arcade learning environment: An evaluation platform for general agents},
  author={Bellemare, Marc G and Naddaf, Yavar and Veness, Joel and Bowling, Michael},
  journal={Journal of artificial intelligence research (JAIR)},
  volume={47},
  pages={253--279},
  year={2013}
}

@article{mnih2015human,
  title={Human-level control through deep reinforcement learning},
  author={Mnih, Volodymyr and Kavukcuoglu, Koray and Silver, David and Rusu, Andrei A and Veness, Joel and Bellemare, Marc G and Graves, Alex and Riedmiller, Martin and Fidjeland, Andreas K and Ostrovski, Georg and others},
  journal={Nature},
  volume={518},
  number={7540},
  pages={529--533},
  year={2015},
  publisher={Nature Publishing Group}
}

@inproceedings{kempka2016vizdoom,
  title={Vizdoom: A doom-based ai research platform for visual reinforcement learning},
  author={Kempka, Micha{\l} and Wydmuch, Marek and Runc, Grzegorz and Toczek, Jakub and Ja{\'s}kowski, Wojciech},
  booktitle={IEEE conference on computational intelligence and games (CIG)},
  pages={1--8},
  year={2016},
  organization={IEEE}
}

@article{vinyals2017starcraft,
author = {Vinyals, Oriol and Ewalds, Timo and Bartunov, Sergey and Georgiev, Petko and Vezhnevets, Alexander and Yeo, Michelle and Makhzani, Alireza and Küttler, Heinrich and Agapiou, John and Schrittwieser, Julian and Quan, John and Gaffney, Stephen and Petersen, Stig and Simonyan, Karen and Schaul, Tom and Van Hasselt, Hado and Silver, David and Lillicrap, Timothy and Calderone, Kevin and Tsing, Rodney},
year = {2017},
month = {08},
title = {StarCraft II: A New Challenge for Reinforcement Learning},
journal = {arXiv:1708.04782}
}

@article{kanervisto2022minerl,
  title={Minerl diamond 2021 competition: Overview, results, and lessons learned},
  author={Kanervisto, Anssi and Milani, Stephanie and Ramanauskas, Karolis and Topin, Nicholay and Lin, Zichuan and Li, Junyou and Shi, Jianing and Ye, Deheng and Fu, Qiang and Yang, Wei and others},
  journal={Competitions and Demonstrations Track (NeurIPS)},
  pages={13--28},
  year={2022},
  publisher={PMLR}
}

@inproceedings{milani2020retrospective,
  title={Retrospective analysis of the 2019 MineRL competition on sample efficient reinforcement learning},
  author={Milani, Stephanie and Topin, Nicholay and Houghton, Brandon and Guss, William H and Mohanty, Sharada P and Nakata, Keisuke and Vinyals, Oriol and Kuno, Noboru Sean},
  booktitle={competition and demonstration track (NeurIPS)},
  pages={203--214},
  year={2020},
  organization={PMLR}
}

@article{fan2022minedojo,
  title={Minedojo: Building open-ended embodied agents with internet-scale knowledge},
  author={Fan, Linxi and Wang, Guanzhi and Jiang, Yunfan and Mandlekar, Ajay and Yang, Yuncong and Zhu, Haoyi and Tang, Andrew and Huang, De-An and Zhu, Yuke and Anandkumar, Anima},
  journal={Advances in Neural Information Processing Systems (NeurIPS)},
  volume={35},
  pages={18343--18362},
  year={2022}
}

@inproceedings{richter2016playing,
  title={Playing for data: Ground truth from computer games},
  author={Richter, Stephan R and Vineet, Vibhav and Roth, Stefan and Koltun, Vladlen},
  booktitle={European conference on computer vision (ECVA)},
  pages={102--118},
  year={2016},
  organization={Springer}
}

@inproceedings{shotton2011real,
  title={Real-time human pose recognition in parts from single depth images},
  author={Shotton, Jamie and Fitzgibbon, Andrew and Cook, Mat and Sharp, Toby and Finocchio, Mark and Moore, Richard and Kipman, Alex and Blake, Andrew},
  booktitle={IEEE/CVF Conference on Computer Vision and Pattern Recognition (CVPR)},
  pages={1297--1304},
  year={2011},
  organization={Ieee}
}

@inproceedings{sturm2012benchmark,
  title={A benchmark for the evaluation of RGB-D SLAM systems},
  author={Sturm, J{\"u}rgen and Engelhard, Nikolas and Endres, Felix and Burgard, Wolfram and Cremers, Daniel},
  booktitle={IEEE/RSJ international conference on intelligent robots and systems (IROS)},
  pages={573--580},
  year={2012},
  organization={IEEE}
}

@article{burke2009optimising,
  title={Optimising engagement for stroke rehabilitation using serious games},
  author={Burke, James William and McNeill, MDJ and Charles, Darryl K and Morrow, Philip J and Crosbie, Jacqui H and McDonough, Suzanne M},
  journal={The Visual Computer},
  volume={25},
  number={12},
  pages={1085--1099},
  year={2009},
  publisher={Springer}
}

@article{cheng2024open,
  title={Open-television: Teleoperation with immersive active visual feedback},
  author={Cheng, Xuxin and Li, Jialong and Yang, Shiqi and Yang, Ge and Wang, Xiaolong},
  journal={Conference on Robot Learning (CoRL)},
  year={2024}
}

@article{he2024omnih2o,
  title={Omnih2o: Universal and dexterous human-to-humanoid whole-body teleoperation and learning},
  author={He, Tairan and Luo, Zhengyi and He, Xialin and Xiao, Wenli and Zhang, Chong and Zhang, Weinan and Kitani, Kris and Liu, Changliu and Shi, Guanya},
  journal={Conference on Robot Learning (CoRL)},
  year={2024}
}

@inproceedings{he2024learning,
  title={Learning human-to-humanoid real-time whole-body teleoperation},
  author={He, Tairan and Luo, Zhengyi and Xiao, Wenli and Zhang, Chong and Kitani, Kris and Liu, Changliu and Shi, Guanya},
  booktitle={IEEE/RSJ International Conference on Intelligent Robots and Systems (IROS)},
  pages={8944--8951},
  year={2024},
  organization={IEEE}
}

@article{sferrazza2024humanoidbench,
    title={HumanoidBench: Simulated Humanoid Benchmark for Whole-Body Locomotion and Manipulation},
    author={Carmelo Sferrazza and Dun-Ming Huang and Xingyu Lin and Youngwoon Lee and Pieter Abbeel},
    journal={Robotics: Science and Systems (RSS)},
    year={2024}
}

@article{tessler2024maskedmimic,
    author = {Tessler, Chen and Guo, Yunrong and Nabati, Ofir and Chechik, Gal and Peng, Xue Bin},
    title = {MaskedMimic: Unified Physics-Based Character Control Through Masked Motion Inpainting},
    year = {2024},
    journal={ACM Transactions on Graphics (TOG)},
    publisher={ACM New York, NY, USA}
}

@article{shrestha2024generating,
  title={Generating Physically Realistic and Directable Human Motions from Multi-Modal Inputs},
  author={Shrestha, Aayam and Liu, Pan and Ros, German and Yuan, Kai and Fern, Alan},
  journal={European Computer Vision Association (ECCV)},
  year={2024}
}

@article{Kendall1989ShapeSurvey,
  author  = {Kendall, David G.},
  title   = {A Survey of the Statistical Theory of Shape},
  journal = {Statistical Science},
  year    = {1989},
  volume  = {4},
  number  = {2},
  pages   = {87--99}
}

@article{kendall1939problem,
  title={The problem of m rankings},
  author={Kendall, Maurice G and Smith, B Babington},
  journal={The annals of mathematical statistics},
  volume={10},
  number={3},
  pages={275--287},
  year={1939},
  publisher={JSTOR}
}

@article{shrout1979intraclass,
  title={Intraclass correlations: uses in assessing rater reliability.},
  author={Shrout, Patrick E and Fleiss, Joseph L},
  journal={Psychological bulletin},
  volume={86},
  number={2},
  pages={420},
  year={1979},
  publisher={American Psychological Association}
}

@incollection{fisher1970statistical,
  title={Statistical methods for research workers},
  author={Fisher, Ronald Aylmer},
  booktitle={Breakthroughs in statistics: Methodology and distribution},
  pages={66--70},
  year={1970},
  publisher={Springer}
}

@article{jerk,
  title={The coordination of arm movements: an experimentally confirmed mathematical model},
  author={Flash, Tamar and Hogan, Neville},
  journal={Journal of neuroscience (JNR)},
  volume={5},
  number={7},
  pages={1688--1703},
  year={1985},
  publisher={Society for Neuroscience}
}

@inproceedings{winkler2022questsim,
  title={Questsim: Human motion tracking from sparse sensors with simulated avatars},
  author={Winkler, Alexander and Won, Jungdam and Ye, Yuting},
  booktitle={ACM Transactions on Graphics (The proceeding of SIGGRAPH Asia)},
  pages={1--8},
  year={2022}
}

@inproceedings{ahuja2022controllerpose,
  title={Controllerpose: inside-out body capture with VR controller cameras},
  author={Ahuja, Karan and Shen, Vivian and Fang, Cathy Mengying and Riopelle, Nathan and Kong, Andy and Harrison, Chris},
  booktitle={Conference on Human Factors in Computing Systems (CHI)},
  pages={1--13},
  year={2022}
}

@misc{zhang2025openwbt,
      title={Unleashing Humanoid Reaching Potential via Real-world-Ready Skill Space}, 
      author={Zhikai Zhang and Chao Chen and Han Xue and Jilong Wang and Sikai Liang and Yun Liu and Zongzhang Zhang and He Wang and Li Yi},
      year={2025},
      eprint={2505.10918},
      archivePrefix={arXiv},
      primaryClass={cs.RO},
      url={https://arxiv.org/abs/2505.10918}, 
}
}

\clearpage
\setcounter{page}{1}
\appendix
\setcounter{table}{0}
\renewcommand{\thetable}{A\arabic{table}}

\maketitlesupplementary
\section{Implementation Details for Real-Time Streaming of Smoothed gameplay~(Smo)}
\label{sec:asynch_system}
To emulate the online gameplay experience of \emph{Just Dance}, we deploy a streaming mode for Smoothed Gameplay~(Smo). In this mode, a live MoCap module extracts human motion directly from the game screen online and continuously transmits the retargeted robot references to the motion tracking controller. To mitigate delay and throughput limitations induced by a computationally heavy GVHMR inference, we adopt a multi-threaded architecture in which a secondary thread interpolates robot frames produced by the retargeting module.

The main thread executes the entire perception and retargeting pipeline, which consists of the Streaming Module (frame capture), the MoCap Module (GVHMR), and the Retarget Module (GMR). Due to the heavy computation required by GVHMR, this pipeline generates an updated target robot pose roughly every 200ms (5Hz). 

A separate interpolation thread smooths the discrete target poses produced by the main thread and sends them to the robot controller via UDP. Since tracking a 5Hz target pose would result in a jerky motion, this thread maintains a buffer of consecutive keyframes ($f_{k-1}$ and $f_k$). Once two frames are available, the interpolator computes intermediate frames ($f_{0...n}^i$, integer $n$ is the number of interpolation frames that can be set by the user) using Linear Interpolation (LERP) for positions and Spherical Linear Interpolation (SLERP) for rotations. This process introduces a constant delay of roughly 0.2s to maintain consistency but effectively upsamples the 5Hz input into a smooth 10–15Hz trajectory. This upsampled stream is then fed into the robot controller, enabling stable and continuous motion execution.

\section{\revised{Broader Embodiment and Generalization}}
\newcommand{\hoverref}{[15]}
\label{sec:broader}
By design, our framework is agnostic to different robot morphologies. In terms of scoring, the Just Dance system is designed for a global population and evaluates motion across diverse body sizes and proportions, making it largely morphology-agnostic. For retargeting, our pipeline builds on the open-sourced GMR framework, which has been successfully applied to multiple humanoid morphologies, including Booster and H1. 
To further validate generality, we apply our pipeline to a HOVER~\hoverref{}-controlled H1 robot in simulation, where the robot successfully follows the reference dance trajectory. A representative snapshot is shown in Fig.~\ref{fig:rebuttal}.
\begin{figure}[t]
\renewcommand{\thefigure}{B1}
    \centering
    \includegraphics[width=1.02\linewidth]{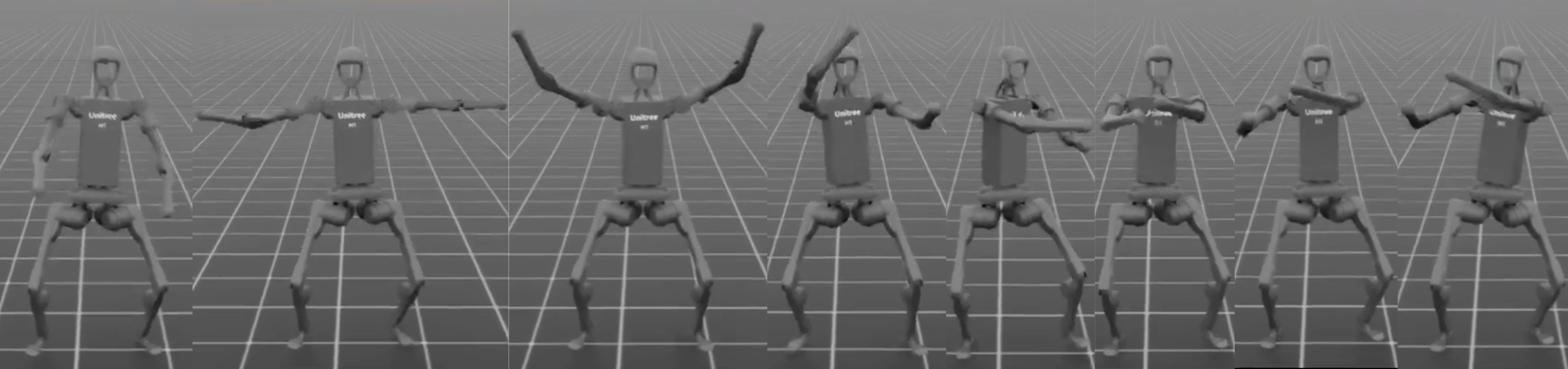}
    \caption{\footnotesize Snapshot of our pipeline being applied to a HOVER~\hoverref{}-controlled H1 robot in simulation, dancing to \textit{Old Town Road; from 0:35 to 0:39.}}
    \label{fig:rebuttal}
\end{figure}

\section{Song Selection for Benchmark Evaluation}
\setcounter{table}{0}
\renewcommand{\thetable}{\thesection\arabic{table}}

For the benchmark, we select songs using two criteria: (1) the absence of hand–floor interaction and (2) minimal in-place turning. Songs that involve dancers placing their hands on the ground are excluded, as the Joy-Con controller is mounted on the robot’s hand, making such interactions infeasible. We additionally exclude songs that require significant in-place turning. Unlike human dancers who wear soft, rounded footwear, the robot’s feet are not designed for continuous pivoting, and these motions demand excessive ankle torque. In practice, we observed that in-place turning gestures are frequently ignored due to hardware limitations.

Applying these criteria, we select five songs for our benchmark, as summarized in Table~\ref{tab:song_list}.

\begin{table}[t]
\centering
\resizebox{\columnwidth}{!}
{
\begin{tabular}{l l l r}
\hline
\textbf{Song} & \textbf{Game Edition} & \textbf{Difficulty } & \textbf{Length (s)} \\
\hline
Old Town Road & JustDance 2020 & Easy (lvl 1) & 161 \\
Heart Of Glass & JustDance 1 & Easy (lvl 2) & 216 \\
Unstoppable & JustDance 2025 Edition & Easy (lvl 2) & 204 \\
Padam Padam & JustDance 2025 Edition & Hard (lvl 3) & 149 \\
Pink Venom & JustDance 2025 Edition & Hard (lvl 3) & 178 \\
\hline
\end{tabular}
}
\caption{Summary of selected Just Dance tracks with game editions, difficulty, and durations. (lvl \#) is the in-game difficulty. we group lvl 1–2 as Easy and 3–4 as Hard.}
\label{tab:song_list}
\end{table}

\section{IRB and Human Subjects Approval}
\label{sec:irb}

This research involves human-motion data extracted from gameplay footage. All procedures involving human data were reviewed and approved by the corresponding Institutional Review Board (IRB). The study was conducted in accordance with applicable ethical guidelines for human subjects research. Consent was obtained where required by the approved protocol.


\end{document}